\documentclass[a4paper,fleqn]{cas-dc}

\usepackage[numbers]{natbib}

\def\tsc#1{\csdef{#1}{\textsc{\lowercase{#1}}\xspace}}
\tsc{WGM}
\tsc{QE}
\tsc{EP}
\tsc{PMS}
\tsc{BEC}
\tsc{DE}

\begin{document}
\let\WriteBookmarks\relax
\def\floatpagepagefraction{1}
\def\textpagefraction{.001}
\shorttitle{RGE-GCN Framework for RNA-seq Cancer Detection}
\shortauthors{Shreyas Shende et~al.}

\title [mode = title]{RGE-GCN: Recursive Gene Elimination with Graph Convolutional Networks for RNA-seq based Early Cancer Detection}                      



\author[1]{Shreyas Shende}

\affiliation[1]{organization={Department of Computer Science, New Jersey Institute of Technology},
                city={Newark},
                postcode={07102}, 
                state={NJ},
                country={USA}}
\credit{Data curation, Writing - Original draft preparation}

\author[1]{Varsha Narayanan}

\author[2]{Vishal Fenn}

\affiliation[2]{organization={Department of Data Science, New Jersey Institute of Technology},
                postcode={07102}, 
                city={Newark},
                state={NJ},
                country={USA}}

\author[2]{Yiran Huang}


\author[3]{Dincer Goksuluk}
\affiliation[3]{organization={Department of Biostatistics},
                city={Sakarya},
                postcode={54187}, 
                country={Turkiye}}

\author[4, 5, 6]{Gaurav Choudhary}
\affiliation[4]{organization={Division of Cardiology, Brown University Health},
                city={Providence},
                postcode={02903}, 
                state={RI}, 
                country={USA}}

\affiliation[5]{organization={Department of Medicine, Warren Alpert Medical School of Brown University},
                city={Providence},
                state={RI},
                postcode={02903},
                country={USA}}

\affiliation[6]{organization={VA Providence Healthcare System},
                city={Providence},
                state={RI},
                postcode={02903},
                country={USA}}

\author[4,5]{Melih Agraz}
 
\author[1,2]{Mengjia Xu}
\ead{mx6@njit.edu}

\cortext[1]{Corresponding author}



\begin{abstract}
Early detection of cancer plays a key role in improving survival rates, but identifying reliable biomarkers from RNA-seq data is still a major challenge. The data are high-dimensional, and conventional statistical methods often fail to capture the complex relationships between genes. In this study, we introduce RGE-GCN (Recursive Gene Elimination with Graph Convolutional Networks), a framework that combines feature selection and classification in a single pipeline. Our approach builds a graph from gene expression profiles, uses a Graph Convolutional Network to classify cancer versus normal samples, and applies Integrated Gradients to highlight the most informative genes. By recursively removing less relevant genes, the model converges to a compact set of biomarkers that are both interpretable and predictive. We evaluated RGE-GCN on synthetic data as well as real-world RNA-seq cohorts of lung, kidney, and cervical cancers. Across all datasets, the method consistently achieved higher accuracy and F1-scores than standard tools such as DESeq2, edgeR, and limma-voom. Importantly, the selected genes aligned with well-known cancer pathways including PI3K–AKT, MAPK, SUMOylation, and immune regulation. These results suggest that RGE-GCN shows promise as a generalizable approach for RNA-seq based early cancer detection and biomarker discovery ({\url{https://rce-gcn.streamlit.app/}}).

\end{abstract}


\date{}

\maketitle

\section{Introduction}
Genomic data analytics has become increasingly critical in advancing our understanding of cancer, particularly in detecting the disease at an early stage~\cite{fitzgerald2022future}.
RNA sequencing (RNA-seq) enables high-resolution examination of gene expression profiles across diverse samples, making it a powerful tool for biomarker discovery. However, the inherently high dimensionality of RNA-seq data involving simultaneous measurements of tens of thousands of genes presents significant computational and statistical challenges. Accurately identifying differentially expressed genes (DEGs), which show meaningful differences between healthy and cancerous samples, is therefore of critical importance. Effective DEG selection not only reduces complexity by highlighting biologically relevant genes but also enhances model interpretability and strengthens classification performance. Ultimately, accurate DEG identification directly contributes to the discovery of reliable biomarkers for early cancer detection, thereby improving patient outcomes and advancing early diagnostic tools.

High-dimensional gene expression data, characterized by a large number of features and a limited number of samples, increase the risk of overfitting and exacerbate computational complexity in classical machine learning approaches \cite{shin2024effective}. To address these limitations, Graph Neural Networks (GNNs) have emerged as powerful tools, offering a structured framework to capture both co-expression patterns and biological interactions. Recent efforts can be broadly grouped into two categories: (1) graph-based models and (2) specialized architectures designed for targeted biomedical applications. (1) Graph-based models. Within the graph domain, comparative studies such as Alharbi et al.~\cite{alharbi2025comparative} evaluated GCN, GAT, and GTN architectures for multi-omics cancer classification, demonstrating the benefit of regularized feature reduction. Similarly, Li and Nabavi~\cite{li2024multimodal} proposed a heterogeneous GNN to integrate inter and intra-omics relationships. However, these approaches are largely centered on multi-omics integration and often depend on preselected feature sets, which constrains their applicability when learning directly from single-omics RNA-seq data. In addition, Wang et al.  \cite{wang2021scgnn} introduced scGNN, a graph-based framework for single-cell transcriptomics that effectively models gene–gene and cell–cell dependencies, while Mao et al.~\cite{bai2024grnnlink} framed gene regulatory networks (GRN) reconstruction as a link-prediction task (GNNLink), using a GCN-based encoder over prior TF–gene graphs and scRNA-seq features to recover regulatory edges; across seven scRNA-seq datasets and multiple ground truths, GNNLink reported competitive or superior AUROC/AUPRC and markedly lower runtime versus several deep baselines. Additionally, Ciortan et al. \cite{ciortan2022gnn} proposed $graph-sc$ models scRNA-seq as a gene-to-cell graph and uses a graph autoencoder to learn cell embeddings for clustering; across 24 simulated and 15 real datasets, it reports competitive ARI/NMI, faster runtime than comparable neural methods, robustness to down-sampling, and easy integration of external gene network. (2) specialized architectures designed for targeted biomedical applications. Cancer focused architectures, such as MSL-GAT for bladder cancer~\cite{ibrahim2025efficient}, illustrate this trend by tailoring models to specific diseases, thereby limiting generalizability. Other pipelines have also shown promise beyond the graph centric paradigm for instance, ML-GAP by Agraz et al.~\cite{agraz2024ml} combined PCA/DEG filtering with autoencoders to achieve strong classification performance. In the single-cell domain, methods such as scGSL~\cite{huang2025graph}, Cellograph~\cite{shahir2024cellograph}, and scDGAE~\cite{feng2023single} address unique challenges like sparsity and cell–cell interaction modeling. Additionally, Qiu et al.~\cite{qiu2021gated} introduced a Gated Graph Attention Network with statistical preprocessing, while Vaida et al.~\cite{vaida2025m} developed a hybrid model combining metabolomics and demographic data for lung cancer detection. GNN have also shown strong potential beyond genomics, particularly in neurological disease classification. Cao et al.\cite{cao2024dementia} introduced a directed structure-learning GNN model that integrates effective brain connectivity features from EEG signals, achieving high accuracy in distinguishing Alzheimer’s disease, Parkinson’s disease, and healthy controls. Their results demonstrate the broader effectiveness of GNN-based medical diagnostic systems.

Despite rapid progress in GNN-based methods for cancer biomarker discovery, most approaches remain tailored to multi-omics integration, cancer-specific pipelines, or single-cell applications. While these frameworks are powerful, they either depend on auxiliary data modalities or rely on preselected feature sets, which limits their applicability in general-purpose workflows. Consequently, there is a notable lack of a generalizable, end-to-end framework that directly operates on bulk RNA-seq data to jointly perform gene selection and classification.

To address this gap, we propose RGE-GCN, a Recursive Gene Elimination Graph Convolutional Network that jointly performs feature selection and sample classification for RNA-seq datasets. While demonstrated on RNA-seq data, the framework is generalizable and can potentially be applied to other omics modalities. Unlike previous graph-based RNA-seq models that depend on preselected DEGs or external networks, RGE-GCN learns intrinsic gene–gene relationships directly from data, yielding both interpretability and scalability. Our approach first constructs a sample–sample graph using the Pearson Correlation Coefficient (PCC) to capture robust co-expression patterns. A Graph Convolutional Network (GCN) is then trained on the complete feature set, and Integrated Gradients (IG) are employed to compute attribution scores for individual genes. Based on these scores, a recursive elimination strategy greedily removes the least informative genes, with the model retrained after each pruning step until a compact yet predictive subset is obtained. This process not only reduces dimensionality but also enhances interpretability by highlighting biologically relevant features. Experiments on both synthetic datasets and multiple public RNA-seq cancer cohorts demonstrate that RGE-GCN achieves competitive or superior classification performance while significantly reducing input dimensionality, thereby enabling scalable and interpretable biomarker discovery.

\section{Materials and methods}

This section details the proposed computational pipeline for identifying minimal, high-performing gene signatures from different RNA-seq datasets. Our framework, illustrated in Fig.~\ref{fig:rge_pipeline}, integrates a GCN within a recursive gene elimination (RGE) procedure. The core of the proposed method, RGE-GCN, lies in constructing a sample-centric graph to model inter-subject relationships and leveraging an axiom-based interpretability method, Integrated Gradients, to guide the feature selection process.  

\subsection{Recursive Gene Elimination with GCNs and IGs for RNA Sequence-based Disease Prediction}
\label{sec:framework}

\begin{figure*}[!h]
    \centering
    \includegraphics[trim={0 70 0 0},clip, width=\textwidth]{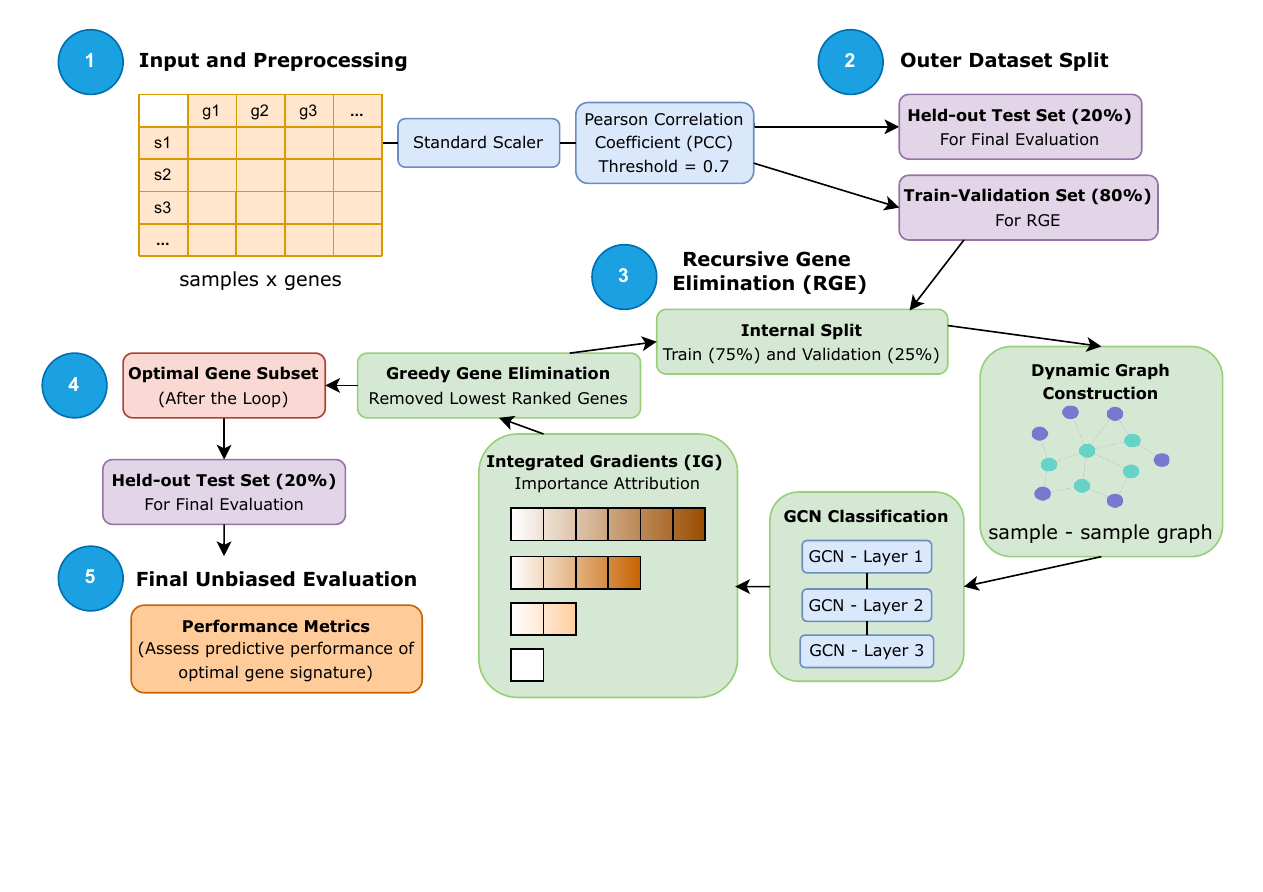}
    \caption{
        \textbf{Architecture of the proposed RGE-GCN framework for biomarker discovery.} 
        The process begins with an outer split of the gene expression data into training and held-out test sets. The main RGE loop operates on the training data, starting with an inner train/validation split. In each iteration, a sample-sample graph is constructed using the Pearson Correlation Coefficient (PCC). A three-layer Graph Convolutional Network (GCN)~\cite{Kipf2016SemiSupervisedCW} is trained on this graph and used to generate class logits. Gene importance scores are then derived from these logits using Integrated Gradients (IG)~\cite{sundararajan2017axiomatic}. Based on these scores, the least informative genes are eliminated. The gene set that maximizes validation accuracy is selected as optimal and is finally evaluated on the held-out test set.
    }
    \label{fig:rge_pipeline}
\end{figure*}

The proposed \textbf{Recursive Gene Elimination with a Graph Convolutional Network (RGE-GCN)} framework offers a comprehensive, end-to-end approach for identifying an optimal gene signature. As the pipeline diagram in Fig.~\ref{fig:rge_pipeline} illustrates, the process begins by first partitioning the dataset into a held-out \textbf{test set} and a combined \textbf{train-validation} set. This design ensures a final, unbiased evaluation of the selected gene signature.

The core of the framework is an iterative, recursive loop that operates on the train-validation set. Each iteration focuses on refining the gene set through a four-stage process:

\begin{itemize}
    \item \textbf{Dynamic Graph Construction:} Within each loop, the train-validation set is further divided into a \textbf{training set} and a \textbf{validation set}. A sample-sample graph is then dynamically constructed from the gene expression profiles of the training data. Here, each node represents a sample, and an edge between two nodes signifies a strong correlation in their gene expression patterns.

    \item \textbf{GCN Classification:} A \textbf{3-layer GCN} is trained on this graph to classify samples based on their disease status. The model’s performance is monitored on the validation set, which serves as a crucial checkpoint for evaluating the quality of the current gene subset.

    \item \textbf{Attributing Gene Importance:} To identify the most influential genes, we employ \textbf{Integrated Gradients}. This technique generates an importance score for every gene, revealing its specific contribution to the GCN’s classification decisions.

    \item \textbf{Greedy Gene Elimination:} Genes are then ranked by their importance scores. Following a greedy strategy, a predefined percentage of the lowest-ranking genes, those with the least impact on classification are eliminated.
\end{itemize}

This recursive cycle continues until a minimum gene count threshold (gene count of the 0.05\%) is met. We have performed an ablation study to select the minimum gene count threshold. An ablation study was conducted to determine this threshold, as detailed in Appendix \ref{sec:ablation}. Throughout the process, the framework keeps track of the gene subset that yields the highest validation accuracy. In the event of a tie, the more concise gene set is selected as the final signature. The chosen signature is then rigorously validated one last time on the initial, unseen held-out test set to confirm its predictive power.

\subsubsection{Greedy Search Nature} The RGE procedure is an instance of a wrapper-based, greedy backward elimination algorithm. It is "greedy" because it makes locally optimal decisions at each iteration by removing genes deemed least important for the \textit{current} model, without any mechanism to reconsider these eliminations. While this heuristic does not guarantee a globally optimal solution, an NP-hard problem for feature selection, it provides a computationally tractable and highly effective strategy for dimensionality reduction in complex biological datasets.

\subsubsection{Input Data and Preprocessing}
\label{sec:preprocessing}

The input to our pipeline is a gene expression matrix $\mathbf{X} \in \mathbb{R}^{N \times G}$, where $N$ is the number of samples and $G$ is the number of genes (features), along with a corresponding vector of labels $\mathbf{y} \in \mathbb{Z}_{K}^{N}$ for $K$ classes.

To ensure robust evaluation and prevent information leakage, the dataset is initially partitioned into a training-validation set (80\%). During the RGE loop, the training-validation set is further subdivided into internal training (75\%) and validation (25\%) sets. The optimal gene set is identified using this internal split, while the hold-out test set is used only once for the final, unbiased performance evaluation.

Prior to model training, feature scaling is applied. A `StandardScaler` is fitted on the training data only to transform features to have zero mean and unit variance. The same fitted scaler is then used to transform the validation and test data.

\subsubsection{Sample-Sample Graph Construction}
\label{sec:graph_construction}

To explicitly model the relationships between subjects, we construct a sample-sample graph $\mathcal{G} = (\mathcal{V}, \mathcal{E})$, where the vertex set $\mathcal{V}$ corresponds to the $N'$ samples in the training set. The graph's weighted adjacency matrix $\mathbf{W}$ is defined by the pairwise similarity of sample expression profiles, measured by the Pearson Correlation Coefficient (PCC). For any two samples $i$ and $j$, with corresponding feature vectors $\mathbf{x}_i, \mathbf{x}_j \in \mathbb{R}^{G'}$, the edge weight $w_{ij}$ is computed as:
\begin{equation}
    w_{ij} = \frac{\operatorname{cov}(\mathbf{x}_i, \mathbf{x}_j)}{\sigma_{\mathbf{x}_i} \sigma_{\mathbf{x}_j}},
\end{equation}
where $G'$ is the number of genes remaining in the current iteration. To construct the final unweighted adjacency matrix $\mathbf{A}$ used by the GCN, we apply a hard threshold $\tau$: an edge $(i, j)$ exists in $\mathcal{E}$ if and only if $|w_{ij}| \ge \tau$. For our experiments, we set $\tau = 0.7$ to ensure the graph captures only strong expression profile similarities.

\subsubsection{Graph Convolutional Network for Sample Classification}
\label{sec:gcn_model}

We employ a Graph Convolutional Network (GCN)~\cite{Kipf2016SemiSupervisedCW} to perform graph-based classification. Each node (sample) is initialized with its corresponding gene expression vector as its feature representation. The GCN updates node embeddings $\mathbf{H}^{(\ell)}$ at each layer $\ell$ according to the propagation rule:
\begin{equation}
    \mathbf{H}^{(\ell+1)} = \sigma\left(\hat{\mathbf{D}}^{-\frac{1}{2}} \hat{\mathbf{A}} \hat{\mathbf{D}}^{-\frac{1}{2}} \mathbf{H}^{(\ell)} \mathbf{W}^{(\ell)}\right),
\end{equation}
where $\hat{\mathbf{A}} = \mathbf{A} + \mathbf{I}$ is the adjacency matrix with self-loops, $\hat{\mathbf{D}}$ is the corresponding diagonal degree matrix, $\mathbf{W}^{(\ell)}$ is a layer-specific trainable weight matrix, and $\sigma(\cdot)$ is a non-linear activation function. \\
\noindent{\textbf{Implementation and Training Details}}
Our model, as shown in Fig.~\ref{fig:rge_pipeline}, is a three-layer GCN with hidden dimensions $64 \rightarrow 32 \rightarrow 16 \rightarrow K$, where $K$ is the number of classes. To stabilize training and mitigate overfitting, we apply `BatchNorm1d` after the first two GCN layers and `Dropout` with a rate of $p=0.4$ after each hidden layer's activation. The model is trained for 200 epochs using the AdamW optimizer with a learning rate of $0.01$ and weight decay of $1 \times 10^{-3}$. To address class imbalance, we utilize a weighted Cross-Entropy Loss function, where weights are calculated as the inverse of class frequencies in the training set.

\subsection{Integrated Gradients for Gene Importance Attribution}
\label{sec:ig}

To determine which genes to eliminate at each RGE step, we require a robust feature attribution method. We employ Integrated Gradients (IG)~\cite{sundararajan2017axiomatic}, an axiom-based interpretability technique that assigns an importance score to each input feature for a trained model's prediction. IG computes the integral of gradients along a linear path from a baseline input $\mathbf{x}'$ to the actual input $\mathbf{x}$. The importance of the $i$-th feature is given by:
\begin{equation}
    \text{IG}_{i}(\mathbf{x}) = (x_i - x'_i) \times \int_{\alpha=0}^{1} \frac{\partial F(\mathbf{x}' + \alpha(\mathbf{x} - \mathbf{x}'))}{\partial x_i} \, d\alpha,
\end{equation}
where $F(\cdot)$ represents the GCN's output function for a specific class. $\alpha \in [0,1]$ denotes the interpolation step from the baseline to the input. 

In our implementation, we use a zero-vector as the baseline $\mathbf{x}'$, representing the absence of any expression signal. The integral is approximated numerically using a Riemann sum with 50 steps. To derive a single, class-agnostic importance score for each gene, we compute its IG value for each of the $K$ output classes and sum their absolute magnitudes. This aggregated score reflects a gene's total influence on the model's decision-making, which is then used for ranking and elimination.
\subsection{Differential Expression Analysis Using Traditional Methods}
\label{sec:deg_analysis}
For comparing our framework against traditional methods in Table \ref{tab:real_results_acc}, \ref{tab:real_results_f1}, we performed differential expression analysis using three commonly used statistical methods: DESeq2, edgeR, and limma-voom. For DESeq2, raw count data was provided and we performed median-of-ratios normalization and dispersion estimation before applying a negative binomial generalized linear model for differential expression testing. For edgeR, the count data was first normalized using the trimmed mean of M-values (TMM) method, followed by dispersion estimation and a generalized linear model likelihood ratio test. For limma-voom, after applying TMM normalization, voom transformation was used to estimate the mean-variance relationship and convert counts to log2-counts-per-million with associated precision weights, enabling linear modeling via empirical Bayes moderation. For classifying genes as differentially expressed, across all three methods two criteria needed to be satisfied: an adjusted p-value (Benjamini-Hochberg FDR correction) less than 0.05 and an absolute log2-fold change greater than 1. These thresholds ensure both statistical significance and biological relevance, filtering for genes with at least a two-fold expression change between conditions.

\subsection{Dataset Description}
\label{sec:data:sets}

We evaluated the proposed method on both synthetic and real-world datasets. A summary of these datasets is provided in Table~\ref{tab:datasets}.

\noindent
\textbf{Synthetic cohorts.} To rigorously benchmark our framework under controlled yet biologically meaningful conditions, we generated de novo RNA-seq–like data using a negative binomial (NB) distribution. Unlike prior studies that directly assign DEG labels, we synthesize \textit{raw count data} by explicitly modeling biological variability, expression intensity, and differential expression through realistic distributional assumptions.This strategy ensures that both expression dynamics and class labels emerge organically from the generative process. Specifically, the raw read count $X_{ij}$ for gene $i$ in sample $j$ was drawn from a negative binomial distribution \cite{goksuluk2023classification}:
\begin{equation}
X_{ij} \sim \mathrm{NB}(\mu_{ij}, \phi_i), \qquad 
\mu_{ij} = s_j \cdot g_i \cdot d_{ik}.
\end{equation}

where $\mu_{ij}$ denotes the expected expression level of gene $i$ in sample $j$. The term $\phi_i$ is the gene-specific overdispersion parameter, with values chosen to represent high overdispersion ($\phi = 0.01$) and low overdispersion ($\phi = 1$). Under the negative binomial parameterization used here, the variance takes the form:
\begin{equation} 
\mathrm{Var}(X_{ij}) = \mu_{ij} + \frac{\mu_{ij}^2}{\phi_i},
\end{equation}

implying that smaller values of $\phi_i$ yield greater dispersion. The sample-specific size factor is given by $s_j \sim \mathrm{Uniform}(0.2,\, 2.2)$, while $g_i \sim \mathrm{Exponential}(25)$ represents the baseline expression level of gene $i$. The differential expression factor $d_{ik}$ (i.e., log fold-change) is drawn from a log-normal distribution  with mean 0 and standard deviation 1 for differentially expressed (DE) genes and set to 1 otherwise.

We constructed multiple cohorts, each comprising 1000 genes measured across 50, 200, or 500 samples. Within these cohorts, the proportion of DEGs varied between 5\% and 30\%. Simulations were conducted under both high and low dispersion scenarios to reflect diverse levels of biological variance \cite{goksuluk2019mlseq}.

The resulting raw count data were preprocessed using a pipeline analogous to established bioinformatics workflows. This included median ratio normalization in the style of DESeq2, followed by a variance stabilizing transformation (VST) and near-zero variance filtering. This comprehensive data generation and preprocessing pipeline enables a reliable and reproducible benchmark for gene selection methods, allowing for robust evaluation against ground-truth labels using metrics such as Accuracy, True Positives (TP), and F1-score across diverse data regimes.

\noindent
{\bf Real-world RNA-seq cohorts.}
Following validation on synthetic datasets, we applied the proposed method to three real-world RNA-seq cohorts:

\begin{itemize}
  \item \textbf{Cervical cancer}  
    58 paired tumour/control samples (714 miRNAs) sequenced on the
    Solexa/Illumina platform by Witten \emph{et al.}~\cite{witten2010ultra}. 
  \item \textbf{Renal cell carcinoma (RCC)}  
    sequencing reads of 20,531 known human RNAs belonging to 1020 primary tumours 1020 primary tumours downloaded from The Cancer Genome Atlas
    (TCGA) portal~\cite{saleem2013linked}. 
    Samples are stratified into three subtypes: KIRP ($n=606$), KIRC ($n=323$),
    and KICH ($n=91$) following \cite{yu2020can} 
  \item \textbf{Lung cancer}  
    Sequencing reads of 20,531 known human RNAs belonging to 1128 tumours (LUAD $=576$, LUSC $=552$) with matched 20 531-gene
    expression counts, also obtained from TCGA~\cite{saleem2013linked}.
\end{itemize}

\begin{table*}[htb!]
\centering
\caption{Summary of the $12$ synthetic and $3$ real-world RNA-seq datasets analyzed in this study. The synthetic RNA-seq data was generated using a Negative Binomial (NB) distribution. In this process, the ground truth (GT) for Differentially Expressed Genes (DEGs) was established by assigning this status to genes based on a fold change sampled from a log-normal distribution. }
\begin{tabular}{lcccc}
\toprule
\textbf{Dataset} & \textbf{\#Samples} & \textbf{\#Genes} & \textbf{Task Type} & \textbf{Disease Context} \\
\midrule
Synthetic cohorts   & 50--500  & 1000   & Cancer vs.~Normal        & Simulated \\
Cervical cancer     & 58       & 714    & Cancer vs.~Normal        & Cancer \\
Kidney cancer (RCC) & 1020     & 20,531 & Cancer subtype (3-class) & Cancer \\
Lung cancer         & 1128     & 20,531 & Cancer subtype (2-class) & Cancer \\
\bottomrule
\end{tabular}
\label{tab:datasets}
\end{table*}

\section{Evaluation Metrics}

To assess how the framework performs both in detecting gene-level DEGs and in 
classifying cancer samples, we conducted experiments using both synthetic and 
real datasets. For each run, we recorded a comprehensive set of evaluation metrics and summarized the results as mean values along with their standard deviations. These metrics correspond directly to those reported in  Table \ref{tab:synth_results_full}-\ref{tab:real_results_f1}, as detailed below.

\begin{itemize}
\item \textbf{Accuracy}: This metric quantifies the proportion of correctly classified samples relative to the total number of samples.
It provides an overall measure of predictive reliability and is reported as the primary performance indicator in Table \ref{tab:real_results_acc}.
\item \textbf{F1-score}: Given that $\text{Accuracy}$ can be misleading in scenarios involving imbalanced datasets, the F1-score is also reported. This metric represents the harmonic mean of $\text{Precision}$ and $\text{Recall}$ and is particularly informative for evaluating model robustness when both false positives and false negatives hold significant importance.

\item \textbf{Macro F1}: In both binary and multi-class settings (e.g., distinguishing between normal and tumor samples or among different RCC or lung cancer subtypes), we employ \emph{Macro F1}, which averages the F1-score equally across all classes, regardless of their sample sizes, to provide a balanced assessment of model performance even under class imbalance.  
\end{itemize}

\begin{table*}[htb!]
\centering
\scriptsize
\caption{Performance of RGE-GCN, RF, SVM, and MLP methods on synthetic RNA-seq datasets.}
\label{tab:synth_results_full}
\begin{tabular}{ccccccccc}
\toprule
No. & \makecell{ Params \\ ($\phi$, n, DE)} & \makecell{\ GT \\ DEGs} & 
\makecell{ \# Sel. \\ Genes} & \makecell{ \# True \\ DEGs} & 
\makecell{ Acc.} & \makecell{ True \\ Acc.} & 
\makecell{ Macro F1} & \makecell{ True \\ Macro F1} \\ \midrule
\multicolumn{9}{c}{\textbf{RGE-GCN}} \\ \midrule
1 & (1, 50, 0.30) & 297 & 52 & 46 & 1.000 $\pm$ 0.000 & 1.000 $\pm$ 0.000 & 1.000 $\pm$ 0.000 & 1.000 $\pm$ 0.000 \\
2 & (1, 50, 0.05) & 54 & 52 & 20 & 0.947 $\pm$ 0.078 & 0.987 $\pm$ 0.027 & 0.946 $\pm$ 0.079 & 0.986 $\pm$ 0.027 \\
3 & (100, 50, 0.30) & 310 & 93 & 68 & \textbf{0.960 $\pm$ 0.053} & 0.947 $\pm$ 0.078 & \textbf{0.956 $\pm$ 0.060} & 0.946 $\pm$ 0.079 \\
4 & (100, 50, 0.05) & 46 & 52 & 33 & \textbf{0.773 $\pm$ 0.090} & 0.560 $\pm$ 0.080 & \textbf{0.769 $\pm$ 0.092} & 0.485 $\pm$ 0.118 \\
5 & (1, 200, 0.30) & 309 & 52 & 51 & 1.000 $\pm$ 0.000 & 1.000 $\pm$ 0.000 & 1.000 $\pm$ 0.000 & 1.000 $\pm$ 0.000 \\
6 & (1, 200, 0.05) & 58 & 52 & 34 & \textbf{1.000 $\pm$ 0.000} & 0.993 $\pm$ 0.008 & \textbf{1.000 $\pm$ 0.000} & 0.993 $\pm$ 0.008 \\
7 & (100, 200, 0.30) & 300 & 52 & 36 & \textbf{0.970 $\pm$ 0.025} & 0.770 $\pm$ 0.069 & \textbf{0.970 $\pm$ 0.025} & 0.765 $\pm$ 0.071 \\
8 & (100, 200, 0.05) & 37 & 52 & 29 & \textbf{0.947 $\pm$ 0.029} & 0.577 $\pm$ 0.065 & \textbf{0.946 $\pm$ 0.029} & 0.526 $\pm$ 0.078 \\
9 & (1, 500, 0.30) & 270 & 52 & 52 & 1.000 $\pm$ 0.000 & 1.000 $\pm$ 0.000 & 1.000 $\pm$ 0.000 & 1.000 $\pm$ 0.000 \\
10 & (1, 500, 0.05) & 48 & 52 & 33 & 1.000 $\pm$ 0.000 & 1.000 $\pm$ 0.000 & 1.000 $\pm$ 0.000 & 1.000 $\pm$ 0.000 \\
11 & (100, 500, 0.30) & 330 & 76 & 57 & \textbf{0.983 $\pm$ 0.003} & 0.792 $\pm$ 0.072 & \textbf{0.983 $\pm$ 0.003} & 0.787 $\pm$ 0.079 \\
12 & (100, 500, 0.05) & 52 & 69 & 41 & \textbf{0.985 $\pm$ 0.012} & 0.748 $\pm$ 0.033 & \textbf{0.985 $\pm$ 0.012} & 0.737 $\pm$ 0.045 \\ \midrule

\multicolumn{9}{c}{\textbf{RF}} \\ \midrule
1 & (1, 50, 0.30) & 297 & 52 & 46 & 0.987 $\pm$ 0.027 & 1.000 $\pm$ 0.000 & 0.986 $\pm$ 0.027 & 1.000 $\pm$ 0.000 \\
2 & (1, 50, 0.05) & 54 & 52 & 20 & 0.933 $\pm$ 0.042 & 1.000 $\pm$ 0.000 & 0.932 $\pm$ 0.044 & 1.000 $\pm$ 0.000 \\
3 & (100, 50, 0.30) & 310 & 93 & 68 & 0.960 $\pm$ 0.080 & 0.960 $\pm$ 0.080 & 0.952 $\pm$ 0.095 & 0.952 $\pm$ 0.095 \\
4 & (100, 50, 0.05) & 46 & 52 & 33 & 1.000 $\pm$ 0.000 & 1.000 $\pm$ 0.000 & 1.000 $\pm$ 0.000 & 1.000 $\pm$ 0.000 \\
5 & (1, 200, 0.30) & 309 & 52 & 51 & 1.000 $\pm$ 0.000 & 1.000 $\pm$ 0.000 & 1.000 $\pm$ 0.000 & 1.000 $\pm$ 0.000 \\
6 & (1, 200, 0.05) & 58 & 52 & 34 & 1.000 $\pm$ 0.000 & 1.000 $\pm$ 0.000 & 1.000 $\pm$ 0.000 & 1.000 $\pm$ 0.000 \\
7 & (100, 200, 0.30) & 300 & 52 & 36 & 1.000 $\pm$ 0.000 & 1.000 $\pm$ 0.000 & 1.000 $\pm$ 0.000 & 1.000 $\pm$ 0.000 \\
8 & (100, 200, 0.05) & 37 & 52 & 29 & \textbf{1.000 $\pm$ 0.000} & 0.997 $\pm$ 0.007 & \textbf{1.000 $\pm$ 0.000} & 0.997 $\pm$ 0.007 \\
9 & (1, 500, 0.30) & 270 & 52 & 52 & 1.000 $\pm$ 0.000 & 1.000 $\pm$ 0.000 & 1.000 $\pm$ 0.000 & 1.000 $\pm$ 0.000 \\
10 & (1, 500, 0.05) & 48 & 52 & 33 & 1.000 $\pm$ 0.000 & 1.000 $\pm$ 0.000 & 1.000 $\pm$ 0.000 & 1.000 $\pm$ 0.000 \\
11 & (100, 500, 0.30) & 330 & 76 & 57 & 1.000 $\pm$ 0.000 & 1.000 $\pm$ 0.000 & 1.000 $\pm$ 0.000 & 1.000 $\pm$ 0.000 \\
12 & (100, 500, 0.05) & 52 & 69 & 41 & 1.000 $\pm$ 0.000 & 1.000 $\pm$ 0.000 & 1.000 $\pm$ 0.000 & 1.000 $\pm$ 0.000 \\ \midrule
\multicolumn{9}{c}{\textbf{SVM}} \\ \midrule
1 & (1, 50, 0.30) & 297 & 52 & 46 & \textbf{0.973 $\pm$ 0.033} & 0.907 $\pm$ 0.116 & \textbf{0.973 $\pm$ 0.033} & 0.904 $\pm$ 0.120 \\
2 & (1, 50, 0.05) & 54 & 52 & 20 & \textbf{0.933 $\pm$ 0.073} & 0.893 $\pm$ 0.010 & \textbf{0.932 $\pm$ 0.074} & 0.892 $\pm$ 0.101 \\
3 & (100, 50, 0.30) & 310 & 93 & 68 & 0.933 $\pm$ 0.060 & 0.960 $\pm$ 0.000 & 0.926 $\pm$ 0.067 & 0.952 $\pm$ 0.095 \\
4 & (100, 50, 0.05) & 46 & 52 & 33 & 1.000 $\pm$ 0.000 & 1.000 $\pm$ 0.000 & 1.000 $\pm$ 0.000 & 1.000 $\pm$ 0.000 \\
5 & (1, 200, 0.30) & 309 & 52 & 51 & 0.997 $\pm$ 0.007 & 1.000 $\pm$ 0.000 & 0.997 $\pm$ 0.007 & 1.000 $\pm$ 0.000 \\
6 & (1, 200, 0.05) & 58 & 52 & 34 & \textbf{0.993 $\pm$ 0.008} & 0.973 $\pm$ 0.008 & \textbf{0.993 $\pm$ 0.008} & 0.973 $\pm$ 0.008 \\
7 & (100, 200, 0.30) & 300 & 52 & 36 & \textbf{0.997 $\pm$ 0.007} & 0.990 $\pm$ 0.013 & \textbf{0.997 $\pm$ 0.007} & 0.990 $\pm$ 0.014 \\
8 & (100, 200, 0.05) & 37 & 52 & 29 & 0.990 $\pm$ 0.013 & 0.997 $\pm$ 0.007 & 0.990 $\pm$ 0.013 & 0.997 $\pm$ 0.007 \\
9 & (1, 500, 0.30) & 270 & 52 & 52 & 1.000 $\pm$ 0.000 & 1.000 $\pm$ 0.000 & 1.000 $\pm$ 0.000 & 1.000 $\pm$ 0.000 \\
10 & (1, 500, 0.05) & 48 & 52 & 33 & 1.000 $\pm$ 0.000 & 1.000 $\pm$ 0.000 & 1.000 $\pm$ 0.000 & 1.000 $\pm$ 0.000 \\
11 & (100, 500, 0.30) & 330 & 76 & 57 & 1.000 $\pm$ 0.000 & 1.000 $\pm$ 0.000 & 1.000 $\pm$ 0.000 & 1.000 $\pm$ 0.000 \\
12 & (100, 500, 0.05) & 52 & 69 & 41 & 1.000 $\pm$ 0.000 & 1.000 $\pm$ 0.000 & 1.000 $\pm$ 0.000 & 1.000 $\pm$ 0.000 \\\midrule
\multicolumn{9}{c}{\textbf{MLP}} \\ \midrule
1 & (1, 50, 0.30) & 297 & 52 & 46 & 0.760 $\pm$ 0.124 & 0.747 $\pm$ 0.217 & 0.732 $\pm$ 0.159 & 0.730 $\pm$ 0.225 \\
2 & (1, 50, 0.05) & 54 & 52 & 20 & 0.787 $\pm$ 0.078 & 0.747 $\pm$ 0.115 & 0.783 $\pm$ 0.078 & 0.726 $\pm$ 0.128 \\
3 & (100, 50, 0.30) & 310 & 93 & 68 & 0.827 $\pm$ 0.116 & 0.827 $\pm$ 0.124 & 0.820 $\pm$ 0.117 & 0.798 $\pm$ 0.144 \\
4 & (100, 50, 0.05) & 46 & 52 & 33 & \textbf{0.947 $\pm$ 0.050} & 0.907 $\pm$ 0.068 & \textbf{0.942 $\pm$ 0.056} & 0.898 $\pm$ 0.080 \\
5 & (1, 200, 0.30) & 309 & 52 & 51 & \textbf{0.963 $\pm$ 0.032} & 0.960 $\pm$ 0.025 & \textbf{0.963 $\pm$ 0.033} & 0.960 $\pm$ 0.025 \\
6 & (1, 200, 0.05) & 58 & 52 & 34 & 0.910 $\pm$ 0.056 & 0.963 $\pm$ 0.036 & 0.909 $\pm$ 0.057 & 0.963 $\pm$ 0.037 \\
7 & (100, 200, 0.30) & 300 & 52 & 36 & 0.967 $\pm$ 0.028 & 0.977 $\pm$ 0.017 & 0.966 $\pm$ 0.028 & 0.976 $\pm$ 0.017 \\
8 & (100, 200, 0.05) & 37 & 52 & 29 & 0.970 $\pm$ 0.032 & 0.973 $\pm$ 0.013 & 0.970 $\pm$ 0.033 & 0.973 $\pm$ 0.014 \\
9 & (1, 500, 0.30) & 270 & 52 & 52 & \textbf{0.992 $\pm$ 0.010} & 0.972 $\pm$ 0.041 & \textbf{0.992 $\pm$ 0.010} & 0.972 $\pm$ 0.042 \\
10 & (1, 500, 0.05) & 48 & 52 & 33 & 0.955 $\pm$ 0.018 & 0.983 $\pm$ 0.022 & 0.954 $\pm$ 0.018 & 0.983 $\pm$ 0.022 \\
11 & (100, 500, 0.30) & 330 & 76 & 57 & 0.988 $\pm$ 0.015 & 0.993 $\pm$ 0.008 & 0.988 $\pm$ 0.015 & 0.992 $\pm$ 0.008 \\
12 & (100, 500, 0.05) & 52 & 69 & 41 & 0.967 $\pm$ 0.023 & 0.976 $\pm$ 0.017 & 0.967 $\pm$ 0.023 & 0.976 $\pm$ 0.017 \\ \bottomrule
\end{tabular}
\begin{flushleft}
\scriptsize
GT DEGs: Ground truth DEGs represents the genes that are labeled differentially expressed within data generation procedure; \# of Sel.Genes: Number of selected genes, which were obtained using Integrated Gradients; \# of True DEGs; Number of True DEGs, which is an intersection between selected genes and Ground Truth; Acc.: Accuracy represents the performance of classification using selected genes; True Acc.:True  Accuracy uses the Ground Truth (all the genes). This table demonstrated the improved performance for each of the models when using Gene Selection. Macro F1: The Macro F1-score is computed by averaging the F1-scores across all classes, giving equal weight to each class regardless of class imbalance. True Macro F1: This is the Macro F1-score obtained when using the full ground-truth DEG set rather than the subset selected by our model, providing an upper-bound reference for classifier performance. 
\end{flushleft}
\end{table*}

\section{Results}

\label{sec:results}

\subsection{Performance on Synthetic Datasets}
\label{sec:results:synth}

 We first evaluated the performance of our proposed RGE-GCN pipeline on a suite of synthetic datasets to assess its robustness under controlled experimental conditions. We made two comparisons:
 1) Using our proposed gene selection, we compare the performance of downstream classification with traditional machine learning models (Random Forest (RF), Support Vector Machine (SVM) \& Multi-Layer Perceptron (MLP)), 2) We used the Ground Truth directly and again compared the performance of downstream classification. The results, summarized in Table \ref{tab:synth_results_full}, demonstrate the high effectiveness of the model in identifying a minimal set of key genes while maintaining excellent classification performance. Across all synthetic cohorts, which were designed to vary in sample size ($n$), overdispersion parameter ($\phi$), and differential expression ratio (DE), the genes selected using RGE consistently achieved near-perfect classification accuracy and F1-scores on most of the models, frequently reaching 1.0000. 

A key observation from these results is the behavior of the gene elimination process. The number of genes predicted by the RGE-GCN consistently converged to a small, highly predictive subset (e.g., 52 genes in most cases), regardless of the considerable variation in the ground-truth number of DEGs (ranging from 37 to 330). This demonstrates the framework's ability to identify a concise, informative gene signature that is sufficient for accurate classification, effectively filtering out extraneous noise. As shown in Table \ref{tab:synth_results_full}, the results from our gene selection are more reliable and consistent than using the Ground Truth, especially under challenging conditions with small sample sizes and low DEG rates, which are common in real-world RNA-seq scenarios. The reason for why we believe the GCN model does not perform as well as SVM and RF on some of the synthetic datasets is due to the synthetic data being too simple, whereas real-world datasets may not follow the same trend.

The performance remained robust across different data regimes, including varying sample sizes and differential expression levels, confirming that the pipeline is well-suited for gene selection tasks in high-dimensional biological data.

\subsection{Performance on Real-world RNA-seq Datasets}
\label{sec:results:real}

Following validation on synthetic cohorts, we benchmarked our proposed pipeline against three widely-used statistical DEG selection methods—DESeq2, edgeR, and limma-voom on three real-world RNA-seq datasets. Each statistical method was used to select a subset of genes, which were then used to train and test a GCN, as well as several traditional machine learning classifiers (RF, SVM, MLP) for comparison. The results, detailed in Table~\ref{tab:real_results_acc}, demonstrate that the proposed RGE-GCN approach consistently identifies a gene subset that yields superior or highly competitive classification accuracy across diverse cancer and disease cohorts.

Across all three datasets, the proposed method combined with a GCN classifier 
achieved the highest or tied-for-highest accuracy, showcasing its effectiveness in 
leveraging gene expression relationships for downstream classification. On the Cervical cancer dataset, the proposed method’s GCN accuracy of 
0.9000 $\pm$ 0.0416 was tied for the highest with limma-voom, and notably outperformed 
results highlight a crucial trade-off. While the statistical methods (DESeq2, edgeR, and 
limma-voom) often selected a smaller number of genes, the RGE-GCN pipeline 
consistently achieved a higher F1-score with its GCN classifier, demonstrating a 
superior balance between precision and recall. 

\begin{table*}[ht!]
\centering
\caption{ Accuracy comparison of the proposed RGE-GCN method against conventional statistical DEG analysis methods (DESeq2, edgeR, and limma-voom) across three real-world RNA-seq datasets (Cervical cancer, Lung cancer, and Kidney cancer).}
\label{tab:real_results_acc}
\footnotesize
\begin{tabular}{l l r r r r r}
\toprule
\textbf{Dataset} & \textbf{Method} & \textbf{\#Genes} & \textbf{RF} & \textbf{SVM} & \textbf{MLP} & \textbf{GCN} \\ \midrule
\multirow{4}{*}{Cervical}
& \textbf{Our Method} & 73 & 0.800 $\pm$ 0.057 & \textbf{\textcolor{red}{0.844 $\pm$ 0.054 $\uparrow$}} & 0.689 $\pm$ 0.057 & \textbf{\textcolor{red}{0.900 $\pm$ 0.042 $\uparrow$} } \\ 
& DESeq2 & 202 & 0.844 $\pm$ 0.042 & 0.767 $\pm$ 0.151 & 0.700 $\pm$ 0.103 & 0.833 $\pm$ 0.079 \\ 
& edgeR & 58 & \textcolor{blue}{0.889 $\pm$ 0.050} & 0.789 $\pm$ 0.119 & \textcolor{blue}{0.767 $\pm$ 0.096} & 0.867 $\pm$ 0.057 \\ 
& limma-voom & 58 & 0.867 $\pm$ 0.044 & 0.744 $\pm$ 0.075 & 0.733 $\pm$ 0.155 & \textcolor{blue}{0.900 $\pm$ 0.042} \\ \midrule
\multirow{4}{*}{Lung}
& \textbf{Our Method} & 3085 & 0.929 $\pm$ 0.008 & \textbf{\textcolor{red}{0.935 $\pm$ 0.013 $\uparrow$}} & \textbf{\textcolor{red}{0.939 $\pm$ 0.013 $\uparrow$}} & \textbf{\textcolor{red}{0.942 $\pm$ 0.016 $\uparrow$}} \\ 
& DESeq2 & 3287 & \textcolor{blue}{0.929 $\pm$ 0.007} & 0.885 $\pm$ 0.008 & 0.916 $\pm$ 0.011 & 0.913 $\pm$ 0.017 \\ 
& edgeR & 883 & 0.928 $\pm$ 0.009 & 0.915 $\pm$ 0.009 & 0.927 $\pm$ 0.014 & 0.922 $\pm$ 0.013 \\ 
& limma-voom & 859 & 0.925 $\pm$ 0.007 & 0.925 $\pm$ 0.007 & 0.913 $\pm$ 0.017 & 0.920 $\pm$ 0.011 \\ \midrule
\multirow{4}{*}{Kidney}
& \textbf{Our Method} & 13471 & \textbf{\textcolor{red}{0.941 $\pm$ 0.008 $\uparrow$}} & \textbf{\textcolor{red}{0.934 $\pm$ 0.009 $\uparrow$}} & 0.928 $\pm$ 0.013 & \textbf{\textcolor{red}{0.942 $\pm$ 0.007 $\uparrow$}} \\ 
& DESeq2 & 2328 & 0.931 $\pm$ 0.006 & 0.876 $\pm$ 0.014 & 0.925 $\pm$ 0.009 & 0.892 $\pm$ 0.025 \\ 
& edgeR & 908 & 0.921 $\pm$ 0.006 & 0.921 $\pm$ 0.015 & 0.929 $\pm$ 0.011 & 0.889 $\pm$ 0.025 \\ 
& limma-voom & 939 & 0.916 $\pm$ 0.003 & 0.911 $\pm$ 0.012 & \textcolor{blue}{0.929 $\pm$ 0.008} & 0.886 $\pm$ 0.023 \\ \bottomrule
\end{tabular}
\begin{flushleft}
Each gene selection method was evaluated using multiple classifiers (RF: Random Forest, SVM: Support Vector Machine, MLP: Multilayer Perceptron, GCN: Graph Convolutional Network), and the table reports the mean accuracy $\pm$ standard deviation across five runs.
\end{flushleft}
\end{table*}

This finding suggests that while statistical methods identify a very concise gene list, 
our GCN-guided approach identifies a more comprehensive, albeit larger, set of genes 
that is ultimately more effective for accurate classification in challenging real-world 
settings. This ability to capture a broader range of relevant signals contributes to the 
consistently high performance observed. 

\begin{table*}[ht]
\centering
\caption{F-1 Score and Gene Count Comparison for Proposed and Statistical Methods on Real-world Datasets.} 
\label{tab:real_results_f1}
\footnotesize
\begin{tabular}{l l r r r r r}
\toprule
\textbf{Dataset} & \textbf{Method} & \textbf{\#Genes} & \textbf{RF} & \textbf{SVM} & \textbf{MLP} & \textbf{GCN} \\ \midrule
\multirow{4}{*}{Cervical}
& \textbf{Our Method} & 73 & 0.876 $\pm$ 0.043 & \textbf{\textcolor{red}{0.828 $\pm$ 0.038 $\uparrow$}} & 0.706 $\pm$ 0.089 & \textbf{\textcolor{red}{0.922 $\pm$ 0.045 $\uparrow$}} \\ 
& DESeq2 & 202 & 0.8635 $\pm$ 0.046 & 0.760 $\pm$ 0.152 & 0.610 $\pm$ 0.146 & 0.843 $\pm$ 0.097 \\ 
& edgeR & 58 & \textcolor{blue}{0.877 $\pm$ 0.055} & 0.787 $\pm$ 0.121 & 0.686 $\pm$ 0.092 & 0.787 $\pm$ 0.090 \\ 
& limma-voom & 58 & 0.854 $\pm$ 0.029 & 0.735 $\pm$ 0.076 & \textcolor{blue}{0.758 $\pm$ 0.077} & 0.899 $\pm$ 0.042 \\ \midrule
\multirow{4}{*}{Lung}
& \textbf{Our Method} & 3085 & 0.930 $\pm$ 0.011 & \textbf{\textcolor{red}{0.942 $\pm$ 0.010 $\uparrow$}} & \textbf{\textcolor{red}{0.950 $\pm$ 0.008 $\uparrow$}} & \textbf{\textcolor{red}{0.948 $\pm$ 0.010 $\uparrow$}} \\ 
& DESeq2 & 3287 & 0.929 $\pm$ 0.007 & 0.884 $\pm$ 0.008 & 0.924 $\pm$ 0.010 & 0.914 $\pm$ 0.015 \\ 
& edgeR & 883 & 0.928 $\pm$ 0.006 & 0.914 $\pm$ 0.009 & 0.926 $\pm$ 0.013 & 0.923 $\pm$ 0.015 \\ 
& limma-voom & 859 & \textcolor{blue}{0.930 $\pm$ 0.007} & 0.911 $\pm$ 0.008 & 0.918 $\pm$ 0.010 & 0.920 $\pm$ 0.013 \\ \midrule
\multirow{4}{*}{Kidney}
& \textbf{Our Method} & 13471 & \textbf{\textcolor{red}{0.913 $\pm$ 0.015 $\uparrow$}} & \textbf{\textcolor{red}{0.897 $\pm$ 0.020 $\uparrow$}} & \textbf{\textcolor{red}{0.901 $\pm$ 0.014 $\uparrow$}} & \textbf{\textcolor{red}{0.907 $\pm$ 0.023 $\uparrow$}} \\ 
& DESeq2 & 2328 & 0.898 $\pm$ 0.023 & 0.808 $\pm$ 0.025 & 0.886 $\pm$ 0.034 & 0.843 $\pm$ 0.033 \\ 
& edgeR & 908 & 0.884 $\pm$ 0.013 & 0.879 $\pm$ 0.024 & 0.895 $\pm$ 0.013 & 0.841 $\pm$ 0.041 \\ 
& limma-voom & 939 & 0.884 $\pm$ 0.019 & 0.863 $\pm$ 0.019 & 0.882 $\pm$ 0.005 & 0.834 $\pm$ 0.016 \\ \bottomrule
\end{tabular}
\begin{flushleft}
The table reports the mean F-1 score $\pm$ standard deviation across five random splits for each gene selection method and classifier combination.
\end{flushleft}
\end{table*}

The results from these real-world datasets align with our findings from the synthetic data, confirming that the RGE-GCN pipeline is a robust and powerful tool for identifying a highly predictive gene signature. The method consistently outperforms or matches classical pipelines, suggesting that its greedy gene elimination strategy and GCN-based feature learning provide a more effective way to select and utilize biologically relevant genes for accurate sample classification.

\begin{figure*}[!h]
    \centering
    \includegraphics[trim={0 60 0 0},clip, width=\textwidth]{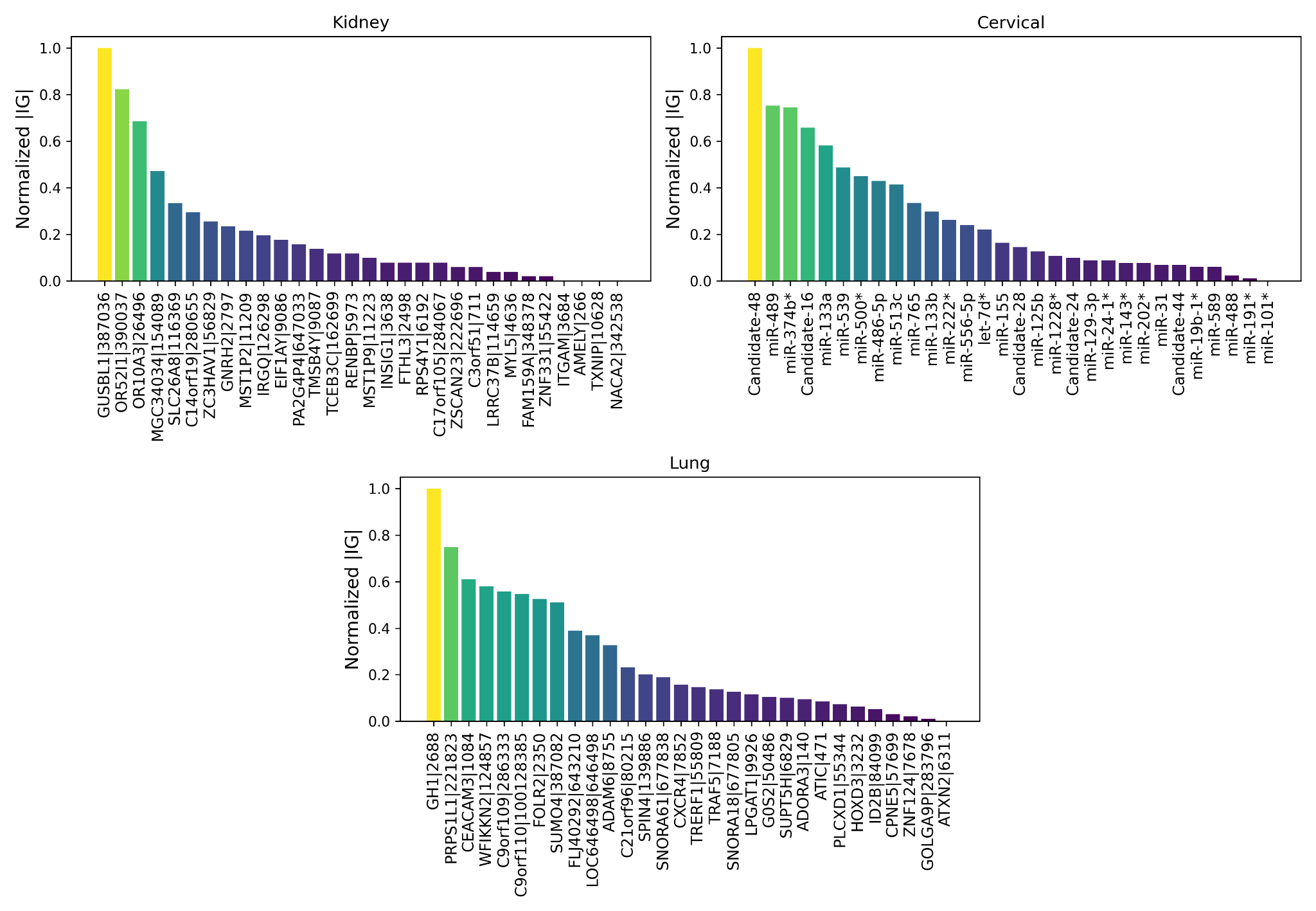}
    \caption{Top 30 genes identified by our RGE-GCN model, ranked according to their normalized Integrated Gradients ($|IG|$) importance scores, for the Kidney, Cervical, and Lung disease datasets.}
    \label{fig:cervical_gcn_importance}
\end{figure*}

\section{Biological validation}

After applying our RGE-GCN framework, which integrates recursive gene elimination with a graph convolutional network and Integrated Gradients, we obtained compact gene sets optimized for classification across multiple cancer cohorts. To assess the biological validity of these selected features, we first carried out a literature-based validation in three cancer types: kidney, lung, and cervical.

\subsection{Scope and gene symbol harmonization}
To assess whether the features selected by our method are biologically meaningful, we carried out a literature-based validation across three cancer types (kidney, lung, and cervical) illustrated in Fig.~\ref{fig:cervical_gcn_importance}. All gene names were harmonized to current HGNC symbols, and noncoding entries (e.g., \textit{miRNA}, \textit{C9orf*}, \textit{FLJ*}, \textit{LOC*}) were treated consistently. For instance, \textit{C9orf109} corresponds to the lncRNA \textit{FAM225A}, \textit{C9orf110} to \textit{FAM225B}, \textit{FLJ40292} is an alias for \textit{EHMT1}, and \textit{ADAM6} is annotated as a human pseudogene. Analyses were conducted at the cohort level without subtype stratification, in line with our experimental design.

\subsubsection{Lung cancer (NSCLC/LUAD).}
Several of the lung cancer features overlap with genes involved in known oncogenic or microenvironmental processes. \textbf{CEACAM3} (CGM1) and \textbf{CEACAM6} belong to the CEACAM adhesion family and show opposite associations with outcome: CEACAM3 positivity correlated with better disease-free survival, while CEACAM6 positivity predicted worse prognosis in EGFR–wildtype LUAD \cite{Kobayashi2012BJC}. \textbf{SUMO4} reflects SUMOylation activity; in NSCLC tissues and models, its upregulation enhanced invasion and migration, whereas knockdown reduced JAK2/STAT3 signaling \cite{Liang2020OncolLett}. \textbf{FOLR2} marks a subset of tumor-associated macrophages; single-cell studies link FOLR2$^{+}$ TAMs to immunosuppressive programs during LUAD progression \cite{Xiang2023CDD}. We also identified olfactory receptors such as \textit{OR52I1} and \textit{OR10A3}, which are increasingly recognized as ectopically expressed GPCRs with functional effects on cancer cell proliferation and migration \cite{Chung2022BMBRep}. In contrast, \textbf{ADAM6} is a pseudogene, suggesting that its appearance in the feature list may reflect linkage or co-expression rather than a driver role (see Discussion).

\subsubsection{Cervical cancer (overall).}
The cervical cancer set was dominated by miRNAs with experimental support in this disease. \textbf{miR-374b} acts as a tumor suppressor through repression of \emph{FOXM1} \cite{xia2019:miR374b}. \textbf{miR-133a} also shows tumor-suppressive activity; its downregulation promotes proliferation, while restoration limits growth via the \emph{LAMB3}–PI3K/AKT axis \cite{sui2023hsa}. By contrast, \textbf{miR-486-5p} is elevated in patient samples and enhances proliferation through PTEN/PI3K–AKT signaling \cite{li2018:miR486}. Together, these findings confirm that the selected miRNAs are linked to major oncogenic pathways central to cervical cancer biology.

\subsubsection{Kidney cancer (overall).}
In kidney cancer, direct evidence was scarcer, but several features suggest potentially relevant mechanisms. Olfactory receptors, also present in our lung list, have been reported across solid tumors with variable roles \cite{Chung2022BMBRep}. Transporters from the SLC superfamily are implicated in metabolic rewiring of clear-cell RCC; here, \textit{SLC26A8}, a testis-enriched anion transporter, lacks specific RCC literature but may represent a novel candidate for further study. Other poorly characterized loci such as \textit{C14orf19} and \textit{MGC34034} may act as regulatory elements rather than direct effectors.

\subsubsection{Gene Ontology (GO) and pathway-level interpretation.}
The lung cancer features mapped to GO processes including \emph{cell–cell adhesion} and \emph{immune regulation} (CEACAMs, FOLR2), \emph{protein SUMOylation} and \emph{signal transduction} (SUMO4–JAK/STAT), and \emph{G protein–coupled receptor signaling} (olfactory receptors) \cite{Kobayashi2012BJC, Liang2020OncolLett, Xiang2023CDD, Chung2022BMBRep}. Cervical cancer miRNAs pointed to \emph{MAPK cascade regulation}, \emph{PI3K signaling}, and \emph{epithelial cell proliferation}, consistent with their validated targets (\emph{FOXM1}, \emph{LAMB3}, \emph{PTEN}). These pathway-level patterns support the biological relevance of the features identified by our method.

\subsubsection{Pathway Analyses} 
The pathway enrichment results in Table \ref{tab:kegg} obtained from the lung cancer cohort revealed that the selected gene set in LUAD clustered along two main axes: (i) metabolic reprogramming focused on nucleotide biosynthesis and redox control, including purine metabolism, folate-mediated one-carbon metabolism, and the pentose phosphate pathway; and (ii) immune regulation within the tumor microenvironment (TME), represented by pathways such as cytokine–cytokine receptor interaction and the intestinal immune network for IgA production.

\begin{table*}[ht]
\caption{KEGG 2021 Human pathway enrichment results for the top 30 genes of lung cancer dataset.}
\centering
\label{tab:kegg}
\begin{tabular}{lccc}
\toprule
\textbf{Name} & \textbf{P-value} & \textbf{Odds Ratio} & \textbf{Combined Score} \\
\midrule
Purine metabolism & 0.01596 & 11.16 & 46.18 \\
One carbon pool by folate & 0.02959 & 36.21 & 127.47 \\
HIV-1 infection & 0.04005 & 6.72 & 21.63 \\
Pentose phosphate pathway & 0.04407 & 23.71 & 74.03 \\
HCMV infection & 0.04459 & 6.33 & 19.67 \\
Endocytosis & 0.05460 & 5.63 & 16.38 \\
Intestinal immune network for IgA & 0.06960 & 14.62 & 38.95 \\
Cytokine-cytokine receptor interaction & 0.07198 & 4.80 & 12.62 \\
Neuroactive ligand-receptor interaction & 0.09229 & 4.14 & 9.86 \\
Small cell lung cancer & 0.1293 & 7.53 & 15.41 \\
\bottomrule
\end{tabular}
\end{table*}

In line with these results, a recent single-cell transcriptomic study demonstrated that purine metabolism is upregulated in LUAD and is associated with poor prognosis and an immunosuppressive tumor microenvironment \cite{zhang2024purine}. Additionally, Yao et al. evaluated five 1CM factors across lung cancer subtypes and found that MTHFD2 and PGDH3 were significantly associated with poor survival only in adenocarcinoma cases \cite{yao2021one}. Essogmo et al. \cite{essogmo2023cytokine} showed that, cytokines were shown to exert both tumor-suppressive and oncogenic roles in lung cancer, occupying a central position in the regulation of immune responses. Moreover, the authors emphasized that this dual functionality has critical implications for immunotherapeutic strategies.

\section{Discussion}
\label{sec:limitations}

In this study, we introduced a new method, the RGE-GCN, designed for RNA-seq–based gene signature discovery. Unlike conventional two-step approaches, our framework integrates gene selection and classification within a single unified process, offering a more streamlined and effective solution. One of the key strengths of RGE-GCN lies in its interpretability, enabled by the use of Integrated Gradients. This feature makes it possible to clearly quantify the contribution of selected genes to model predictions, which in turn facilitates biological validation and supports the generation of new hypotheses. Moreover, proposed framework uses model-based importance scoring for dimensionality reduction, working without predefined statistical thresholds or gene-level annotations. We tested the model on both synthetic and real-world datasets, and it held up well delivering reliable results across multiple cancer types and neurodegenerative disease cohorts. For this reason, we can say that RGE-GCN can be applied to RNA-Seq datasets. Equally important, many of the gene and miRNA signatures uncovered by the model were linked to well-established cancer related pathways such as PI3K–AKT, MAPK, SUMOylation, and immune regulation. These connections, supported by the existing literature, suggest that the method is not only computationally effective but also biologically informative, offering potential for discovering clinically relevant biomarkers.

RGE-GCN framework demonstrates strong ability in identifying compact and predictive gene signatures; however, it can have some limitations when interpreting its results. First, the computational complexity of the recursive elimination process can present a major challenge. At each iteration, the GCN model must be retrained from scratch and Integrated Gradients must be recalculated, which can become particularly costly for large RNA-seq datasets. Future work could address this issue by incorporating transfer learning to initialize subsequent iterations. Alternatively, applying the framework after preliminary dimensionality reduction techniques such as PCA could further improve its efficiency. In addition, the current framework is unimodal, relying solely on gene expression data. However, complex diseases like cancer are driven by perturbations across multiple molecular layers. A significant avenue for future work involves extending the model to integrate multi-omics data. Node features could be enriched with information from DNA methylation, copy number variation (CNV), proteomics, or single-cell ATAC-seq. Developing methods for graph-based fusion of these modalities remains a challenging but promising frontier~\cite{wu2020comprehensive}, with the potential to uncover regulatory programs invisible at the RNA level alone.

When we compared our graph-guided feature selection with the literature, we saw that many of the prioritized signals  well matched with the tumor biology. For instance, in lung cancer, we identified signatures tied to CEACAM-family adhesion molecules, SUMO-pathway activity, and FOLR2 macrophage programs. These findings point to multiple layers of tumor biology. They cover how cells adhere to and interact with the extracellular matrix, how protein signaling is fine-tuned through post-translational modifications, and how the immune microenvironment is remodeled. Taken together, such processes are likely to act in concert, fueling tumor progression and shaping therapeutic response \cite{Kobayashi2012BJC, Liang2020OncolLett, Xiang2023CDD}. We also observed ectopic expression of olfactory receptors, a result that aligns with the growing recognition of GPCR-linked signaling modules in cancer biology \cite{Chung2022BMBRep}. In the case of cervical cancer, several miRNA signals (miR-374b, miR-133a, miR-486-5p, miR-489) repeatedly mapped to PI3K–AKT and MAPK pathways, suggesting that our framework detects functional regulators rather than dataset-specific artifacts. At the same time, not all of the signals are well understood. Genes such as \textit{SLC26A8}, \textit{C14orf19}, and \textit{MGC34034} from the kidney set, or \textit{ADAM6} from the lung set (currently annotated as a pseudogene), do not yet have clear mechanistic roles in human disease. These less-characterized features may represent intriguing directions for further study. We interpret them as potential regulatory markers tagging nearby functional loci, context-dependent features tied to the cell of origin or stromal environment, or entirely novel candidates worthy of deeper exploration. For such cases, we recommend (i) independent expression validation in tumor versus adjacent normal samples (RNA-seq/qPCR), (ii) target validation of miRNA–mRNA interactions via luciferase reporter assays, (iii) functional perturbations (siRNA/ASO/CRISPRi for lncRNAs; siSUMO4; CEACAM3 overexpression) with phenotypic readouts on invasion, migration, and JAK/STAT or PI3K–AKT signaling, and (iv) microenvironmental profiling (IHC/IMC) of FOLR2$^{+}$ TAMs in LUAD cohorts to test associations with stage and outcome. Overall, the overlap between our selected features and published evidence across adhesion, SUMOylation, PI3K–AKT/MAPK signaling, and immune suppression highlights the biological and clinical relevance of the discovered signatures. For genes and loci without existing mechanistic literature, our framework nominates testable hypotheses and provides a foundation for future experimental studies. Additionally, we showed that our findings are corroborated by PubMed-indexed studies. In lung cancer, the highlighted features CEACAM3 \cite{kobayashi2012carcinoembryonic}, GUSB1 \cite{krasnov2011novel}, FOLR2 \cite{xu2018silencing}, and SUMO4 \cite{liang2020sumo4} are well-documented in the literature. Similarly, in cervical cancer,  signatures such as miR-374b \cite{xia2019mir}, miR-500a, miR-486-5p \cite{sharma2016novel}, miR-133b \cite{wang2024roles} and miR-133a \cite{sui2023hsa} converge on key oncogenic pathways. Finally, for kidney cancer, MGC34034 (LINC01187) \cite{mannan2023characterization}, MST1P2 \cite{li2024yap} and GNRH2 \cite{zhang202013} have also been reported as relevant. Together, these literature-backed associations reinforce the biological validity of our graph-guided selection and support the clinical relevance of the identified biomarkers.

\section{Conclusion}
\label{sec:conclusion}





In this study, we introduced a novel end-to-end framework, the Recursive Gene Elimination with Graph Convolutional Networks (RGE-GCN), for biomarker discovery and classification in high-dimensional RNA-seq data. By integrating gene selection directly into the model training pipeline, RGE-GCN circumvents the need for conventional statistical filtering and ground-truth DEG annotations, offering a flexible and data-driven approach. The framework leverages a sample-sample graph to capture inter-sample relationships and employs Integrated Gradients to transparently guide a greedy recursive elimination strategy.

Our empirical evaluation on both synthetic and real-world datasets demonstrates the robust performance and practical utility of the RGE-GCN pipeline. On synthetic cohorts, the method consistently achieved near-perfect classification accuracy and F1-scores, proving its ability to distill a small, highly predictive gene subset from a large, complex feature space. The results on public RNA-seq datasets further validated these findings. In head-to-head comparisons, our proposed method, particularly when utilizing a GCN classifier, consistently achieved superior or competitive accuracy compared to established methods such as DESeq2, edgeR, and limma-voom, most notably on the Kidney cancer datasets.

The core contribution of this work is the development of a unified framework that holistically addresses feature selection and classification. This integrated approach ensures that the selected gene signature is not merely statistically significant but is also optimized for a specific predictive task, leading to more meaningful and actionable biological insights. While the current greedy search and computational complexity present avenues for future work, our findings establish a strong foundation for graph-based, interpretable machine learning in genomics. The RGE-GCN framework represents a significant step toward developing scalable and generalizable tools for discovering minimal yet highly predictive biomarkers, paving the way for improved early-stage disease detection and personalized medicine.

\section{Acknowledgments}

This work was supported by the DOE SEA-CROGS project (DE-SC0023191), AFOSR project (FA9550-24-1-0231). We also acknowledge the computing resources provided by the High Performance Computing (HPC) facility at NJIT.


\appendix
\section{My Appendix}
\subsection{Ablation Study}
\label{sec:ablation}

To understand the sensitivity of the RGE-GCN pipeline to its key hyperparameters and to justify the final model configuration, we conducted an ablation study focusing on two critical parameters: the random seed and the min-genes threshold. The greedy nature of our recursive elimination procedure means that the final gene set and model performance are influenced by the specific path taken during optimization. This study was designed to quantify this variability and establish a robust strategy for selecting the most effective gene signature for our real-world datasets.

Our primary RGE-GCN framework iteratively reduces the gene count until a predefined minimum threshold is reached. We evaluated the framework's performance with three min-genes thresholds: 5\%, 10\%, and 20\% of the initial gene set. The aggregated results, presented in Table \ref{tab:ablation_results_real}, demonstrate that the 5\% threshold generally yielded the most compelling balance between feature reduction and predictive performance across the datasets. The 5\% threshold delivered the highest accuracy and F1-score on Lung and Kidney datasets, confirming its efficacy in distilling key biomarkers. The Cervical dataset showed consistent performance across all thresholds, but with a significantly lower gene count at 5\% and 10\%.

The choice of a random seed also plays a significant role in a greedy search algorithm, as it can influence the initial model weights and, consequently, the gene importance rankings and the recursive elimination path. This variability confirms that there is no single, universally optimal seed for gene elimination. The challenge, therefore, lies not in finding a universal rule but in adopting a pragmatic strategy to consistently select a high-performing result.

Given this inherent challenge, our final strategy for reporting the main results was to select the gene signature that achieved the highest validation accuracy across all iterations and seeds. This pragmatic approach effectively navigates the sub-optimality of the greedy search by identifying the most successful outcome for a given dataset. The ablation study on real-world datasets consistently pointed to the 5\% min-genes threshold as the most effective for achieving high performance with a minimal number of genes. This finding validates our final hyperparameter choices and provides transparency into the RGE-GCN's behavior, reinforcing the credibility of our reported findings.

\begin{table*}[!t]
\centering
\caption{Ablation Study of min-genes Threshold on Real-world Datasets}
\label{tab:ablation_results_real}
\begin{tabular}{lccccc}
\toprule
\textbf{Dataset} & \textbf{Min. Gene Threshold} & \textbf{Best Seed} & \textbf{Gene Count} & \textbf{Avg. Accuracy} & \textbf{Macro F1-Score} \\
\midrule
\multirow{3}{*}{Cervical} & 5\% & 200 & 123 & \textbf{0.9333$\pm$0.0222} & \textbf{0.93} \\ 
& 10\% & 200 & 123 & \textbf{0.9333$\pm$0.0222} & \textbf{0.93} \\ 
& 20\% & 200 & 185 & \textbf{0.9333$\pm$0.0222} & \textbf{0.93} \\ \hline
\multirow{3}{*}{Lung} & 5\% & 300 & 1078 & \textbf{0.9534$\pm$0.0047} & \textbf{0.95} \\  
& 10\% & 200 & 2777 & 0.9499$\pm$0.0100 & 0.95 \\ 
& 20\% & 300 & 5801 & 0.9469$\pm$0.0097 & 0.95 \\ \hline
\multirow{3}{*}{Kidney} & 5\% & 100 & 1477 & \textbf{0.9627$\pm$0.0064} & \textbf{0.94} \\ 
& 10\% & 42 & 2250 & 0.9601$\pm$0.0056 & 0.94 \\ 
& 20\% & 42 & 7956 & 0.9490$\pm$0.0070 & 0.93 \\ 
\bottomrule
\end{tabular}
\end{table*}

\end{document}